# Narrative Feature or Structured Feature? A Study of Large Language Models to Identify Cancer Patients at Risk of Heart Failure

**Ziyi Chen, MS[1], Mengyuan Zhang, BS[1], Mustafa Mohammed Ahmed, MD[2], Yi Guo, PhD[1], Thomas J. George, MD[3], Jiang Bian, PhD[1], Yonghui Wu, PhD[1*]**
**[1]Department of Health Outcomes and Biomedical Informatics, College of Medicine, University of Florida, Gainesville, FL, USA; [2]Division of Cardiovascular Medicine, Department of Medicine, College of Medicine, University of Florida, Gainesville, FL, USA; [3]Division of Hematology & Oncology, Department of Medicine, College of Medicine, University of Florida, Gainesville, FL, USA**

**Abstract**
*Cancer treatments are known to introduce cardiotoxicity, negatively impacting outcomes and survivorship. Identifying cancer patients at risk of heart failure (HF) is critical to improving cancer treatment outcomes and safety. This study examined machine learning (ML) models to identify cancer patients at risk of HF using electronic health records (EHRs), including traditional ML, Time-Aware long short-term memory (T-LSTM), and large language models (LLMs) using novel narrative features derived from the structured medical codes. We identified a cancer cohort of 12,806 patients from the University of Florida Health, diagnosed with lung, breast, and colorectal cancers, among which 1,602 individuals developed HF after cancer. The LLM, GatorTron-3.9B, achieved the best F1 scores, outperforming the traditional support vector machines by 39%, the T-LSTM deep learning model by 7%, and a widely used transformer model, BERT, by 5.6%. The analysis shows that the proposed narrative features remarkably increased feature density and improved performance.*

**Introduction**
Cancer and cardiovascular disease are the top 2 causes of death in the United States (US), and they commonly coexist and intersect at multiple levels.[1–3] Cancer is a significant public health issue globally and the second most common cause of death in the US. In 2023, there are 1,958,310 new cases of cancer in the US, leading to 609,820 deaths.[4] Lung and bronchus cancer is the most fatal form of cancer, responsible for an estimated 127,070 deaths, followed by colorectal cancer, with an estimated 52,550 deaths. Breast cancer is the most common cancer diagnosis, with an estimated 300,000 individuals. Many cancer treatment modalities, such as chemotherapy and radiotherapy, are known to introduce cardiotoxicity and can lead to cardiac malfunction, which is a significant cause of illness and death among cancer patients.[5] Cancer patients are thus often faced with the dual challenge of not only managing their primary cancer but also navigating the potential cardiotoxic effects of cancer treatments.[6] Even if not directly cardiotoxic, cancer treatments can result in changes to metabolism, energy balance, anemia, and other physiologic stressors that may accelerate or uncover pre-existing predispositions to heart disease. To address this issue, cardio-oncology is a growing field of interest in combining the knowledge of cardiology and oncology to identify, observe, and treat cardiovascular disease in cancer patients. The incidence of HF significantly limited treatment options for cancer and negatively impact the quality of life. There is an increasing interest in identifying cancer patients at risk of HF using electronic health records (EHRs) to help decision making and improve the safety of cancer treatments.

Prediction of HF is typically approached as a binary classification task, which is approached using machine learning models to classify a given individual into either positive (at HF risk) or negative (at no HF risk) categories. Previous studies have explored the use of EHRs to predict the risk of HF using traditional machine learning models and deep learning models based on neural networks. Yang *et al.*[7] systematically explored traditional machine learning models, including logistic regression (LR), random forest (RF), support vector machines (SVMs), and gradient boosting(GB) with one-hot and term frequency-inverse document frequency (TF-IDF) feature encoding strategies. Angraal *et al.*[8] developed models to predict mortality and hospitalization for outpatients with HF with preserved ejection fraction (HFpEF) using LR, RF, GB, and SVMs. Yu *et al.*[9] explored genomics data from the UK biobank for heart failure prediction. In these previous studies, the structured medical codes from EHRs are typically represented as a vector with values of zeros and ones, where zeros indicate the patient doesn't have the corresponding feature and ones mean the patient has the corresponding features, known as one-hot encoding. However, during the one-hot encoding, the time-to-event structure of EHRs is simplified to a vector representation without considering the temporal relations. To capture the time-to-event structure, researchers have explored deep learning methods such as the recurrent neural network implemented using long short-term memory (LSTM)[10]. Hybrid neural networks[11–13], including hybrid convolutional neural networks (CNNs) and LSTM method[12], and cluster-based bi-directional LSTM (C-BiLSTM)[13], were explored, demonstrating better performance than traditional machine learning methods. Previous studies have explored Transformer models for cardiovascular disease prediction. For example, Rasmy *et al*.[14,15] applied LSTM and a popular Transformer model, Med-BERT, for HF prediction in diabetic patients using structured EHRs; Antikainen *et al*.[16] compared BERT and XLNet for cardiovascular disease prediction using structured EHRs.

Most recently, transformer models trained using massive amounts of text data with billions of parameters – known as large language models (LLMs) – have demonstrated advanced abilities in many artificial intelligence (AI) tasks, such as natural language processing (NLP), disease prediction, and mortality prediction. Recent studies[17–19] have shown that LLMs are all-purpose prediction engines that can be used for many prediction tasks. A previous study has explored NYUTron[17], a BERT-based transformer model for readmission prediction using only clinical narratives. NYUTron is trained using 110 million parameters, which is relatively small. We previously explored various transformer models[20–22] and developed GatorTron[18], an LLM based on the BERT architecture with up to 8.9 billion parameters and trained using 90 billion words of text.

However, it's unclear how to better use structured EHRs in LLMs for disease risk prediction. Jiang *et al.*[17] used only clinical notes for readmission prediction, which are narrative text. This study aimed to fill this gap by investigating LLMs to identify cancer patients at risk of HF using structured EHRs. We identified patients diagnosed with lung, breast, or colorectal cancer in EHRs from the University of Florida (UF) Health. We then identified HF cases after cancer diagnosis and developed machine learning models to identify cancer patients at risk of HF. This study explored a novel strategy to use structured EHRs as narrative features in LLMs, denoted as "subword features", for HF prediction. The experimental results show that GatorTron models with subword features achieved the best performance, outperforming LSTM and traditional machine learning models. We systematically examined traditional one-hot encoding in traditional machine learning models, sequence representation in LSTM, and subword representation in LLMs. The results show that the subword feature remarkably increased feature overlap among samples, thus benefiting the prediction. We also examined the performance of our models for three cancer subtypes. This study demonstrates the efficiency of using LLMs with subword features for disease onset prediction. Our results show that the subword feature is a good surrogate for using structured medical codes as features in LLMs. The proposed methods can be used to improve cancer treatment outcomes and safety.

## Methods

### Data source

We identified lung, breast, and colorectal cancer patients using International Classification of Disease version 9 (ICD-9) and International Classification of Disease version 10 (ICD-10) codes from the UF Health Integrated Data Repository (IDR) and collected their structured EHRs between 2012 and 2020, including demographics, diagnosis, and medications. This study was approved by UF's Institutional Review Board #202003159.

Cases were defined as cancer patients who had at least one HF diagnosis after the date of cancer diagnosis and had a visit at least six months before HF onset. The onset date of HF was defined as the first HF diagnosis date. We reused a list of ICD 9/10 codes defined by domain experts for heart failure in our previous study[7]. Patients who had HF diagnosis before cancer were excluded. Non-cases were defined as cancer patients without any HF diagnosis. To ensure enough data for prediction, we excluded patients who had less than two visits. We kept the natural ratio between cases and non-cases from this real-world cancer population identified at UF Health.

Table 1 shows the detailed number of cases and non-cases and the distribution for cancer types, age, sex, race, and ethnicity. There are a total number of 12,806 unique cancer patients identified, where 1,602 of them are cases and 11,204 of them are non-cases. The distribution of demographic information is similar between the case and non-cases, except for a slightly higher proportion of cancer cases in the black patients compared with the white group. There are 4,686 patients who only have lung cancer, 4,592 patients only have breast cancer, and 2,718 patients only have colon cancer. There are 810 patients who have multiple types of cancer.

Table 1. Comparison of cases and non-cases.

| Features | Subtypes | Case (n=1,602) | Non-case (n=11,204) | Total (n=12,806) |
|---|---|---|---|---|
| **Cancer Type** n (%) | Only lung cancer | 593 (37.0%) | 4,093 (36.5%) | 4,686 (36.6%) |
| | Only breast cancer | 568 (35.5%) | 4,024 (35.9%) | 4,592 (35.9%) |
| | Only colorectal cancer | 271 (16.9%) | 2,447 (21.8%) | 2,718 (21.2%) |
| | More than one cancer type | 170 (10.6%) | 640 (5.7%) | 810 (6.3%) |
| **Sex** n (%) | Male | 503 (31.4%) | 3,735 (33.3%) | 4,238 (33.1%) |
| | Female | 1,099 (68.6%) | 7,469 (66.7%) | 8,568 (66.9%) |
| **Race** n (%) | White | 1,066 (66.5%) | 8,341 (74.4%) | 9,407 (73.5%) |
| | Black or African American | 476 (29.7%) | 2,055 (18.3%) | 2,531 (19.8%) |
| | Asian | 11 (0.7%) | 163 (1.5%) | 174 (1.4%) |
| | Other | 45 (2.8%) | 550 (4.9%) | 595 (4.6%) |
| | Unknown | 4 (0.2%) | 95 (0.8%) | 99 (0.8%) |
| **Ethnicity** n (%) | Hispanic or Latino | 39 (2.4%) | 366 (3.3%) | 405 (3.2%) |
| | Not Hispanic or Latino | 1,556 (97.1%) | 10,714 (95.6%) | 12,270 (95.8%) |
| | Unknown | 7 (0.4%) | 124 (1.1%) | 131 (1.0%) |
| **Age** | Median [Q1, Q3] | 64.0 [56.0, 73.0] | 61.0 [53.0, 69.0] | 62.0 [53.0, 70.0] |

**Machine learning features**

We extracted patient demographics from EHRs, including gender, race, age, ethnicity, cancer type, smoking status, diagnoses coded in ICD codes, and medications in RxNorm codes. Only the diagnosis and medication records before the HF onset date were used for prediction.

***One-hot encoding for traditional machine learning features***

As the ICD and RxNorm codes are sparse, we used PheWAS groups[23] to aggregate the ICD codes and used ingredient-level RxCUI (RxNorm Concept Unique Identifier) for medications. After grouping, the 23,522 unique ICD-9 and ICD-10 codes were aggregated into 3,706 unique PheWAS groups; the 19,685 clinical-level RxNorm codes were aggregated into 5,578 ingredient-level RxCUI codes. To further reduce sparsity, we removed PheWAS and RxCUI codes that appeared in less than ten patients, resulting in 2,221 unique diagnosis codes and 1,708 unique medication codes. We used the typical one-hot encoding to represent patients in a vector with values of zeros and ones.

***Sequence feature for LSTM***

We organized patient data into sequences according to the encounter time to capture the time-to-event structure of EHRs. Following the previous study, we created mini-batches to group samples with the same sequence length. Each mini-batch has three dimensions: the mini-batch size, the sequence length, and the input dimension. Within the mini-batch, the diagnosis codes and medication codes were organized into a sequence according to the encounter time. We also have labels and elapsed time tensors in the same format. The elapsed time is measured in days. After conversion, there are a total of 417,419 sequences from 12,806 patients, with a minimum sequence length of 2.

***Subword feature for LLMs***

Subwords are word pieces that are generated by the tokenizer of LLMs. LLMs typically use a tokenizer to cut the input words into pieces (i.e., subwords) to increase the overlap of word features. To use structured codes in LLMs, we used the official descriptions of the ICD codes and RxNorm codes and organized them into a text according to their encounter time, where the left to right sequence in the text was used as a surrogate of the encounter time. Figure 1 shows the feature encoding strategies for LSTM and LLMs. Specifically, we converted patient demographics into a narrative description, arranged the diagnosis and medications into a sequence according to their encounter time, kept the unique codes in the last place for each patient, and converted them into official descriptions. Then, we concatenated the two sections into a narrative paragraph as the input. During training, we used the tokenizers of LLMs to break the input into a sequence of subwords and generated a vector representation for classification. Therefore, these features are denoted as subword features. The dimensions of the input data were reduced from 3952 to 4 when the feature encoding method was changed from one-hot encoding to the subword method.

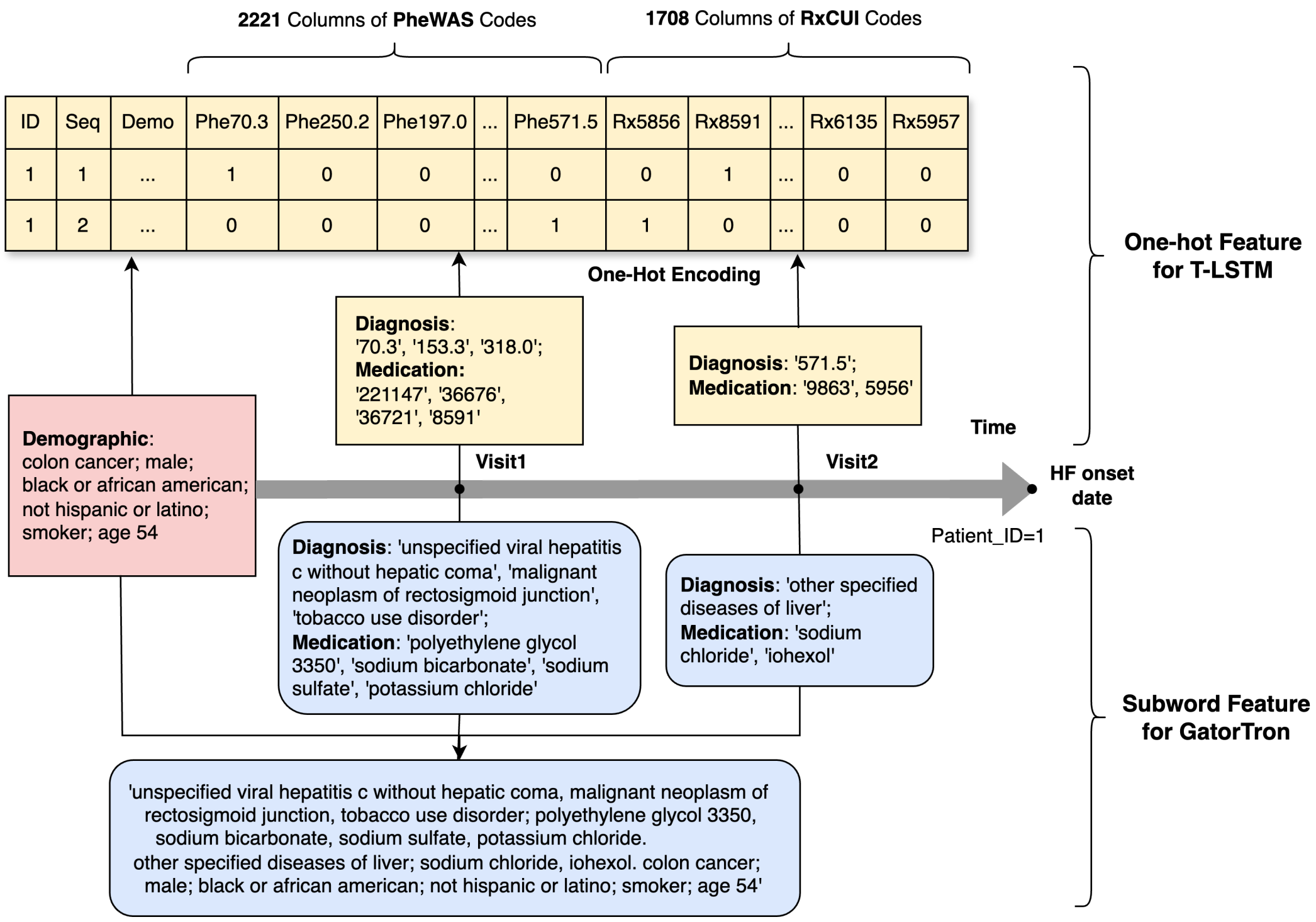

**Figure 1.** An example of generating sequence features for LSTM and subword features for LLMs. Given a patient with two visits, their diagnosis codes were grouped using the PheWAS code, and medications were grouped using the ingredient-level RxCUI code and organized into sequences according to the encounter time. For LLMs, the ICD and RxCUI medication codes are organized into sequences according to the encounter time and then converted into a narrative paragraph using their official descriptions.

## Machine learning models

This study explored traditional machine learning models, deep learning models based on LSTM, and LLMs.

***Traditional machine learning models***. We explored XGBoost (XGB)[24] and support vector machines (SVMs)[25], which are widely used for disease prediction. For SVMs, we adopted implementations that are available in the sklearn.svm library. We optimized model parameters, such as the Radial Basis Function (RBF) kernel, the kernel coefficient (gamma), the regularization parameter c, and the degree of the polynomial kernel function. For XGB, we used the implementation from the xgboost package and optimized the learning rate (eta), the maximum depth of a tree (max_depth), and the number of boost trees (n_estimators).

***LSTM model.*** LSTM is a deep learning model previously used to capture the time-to-event structure of EHRs to improve disease risk prediction. In this study, we used Time-Aware LSTM (T-LSTM)[26], a variant of LSTM designed to handle irregular elapsed times between successive sequential data elements. T-LSTM incorporates elapsed time information into the standard LSTM architecture to capture the temporal dynamics of sequential data with time irregularities. This sequential structure improved the representation of complex historical information about patients with different elapsed periods and sequence lengths.

***Large language models***. We explored BERT[27] and GatorTron. BERT is the first transformer model that succeeded in deep learning and is widely used for perdition tasks. GatorTron is a clinical LLM developed using the BERT architecture and trained with 90 billion unstructured data, such as clinical, biomedical, and general English texts. Within the 90 billion words, there are 80 billion clinical words from over 290 million notes sourced from the UF

Health system. These notes encompass patient records from 2011 to 2021 from over 126 clinical departments and approximately 50 million encounters. GatorTron models have shown superior performance to other transformer models, such as BERT, RoBERTa[28], LongFormer[29], DeBERTa[30], and BioBERT[31], in the biomedical and clinical domains. In this study, we explored the GatorTron-base with 345 million parameters, denoted as GatorTron-345M, and GatorTron-medium with 3.9 billion parameters, denoted as GatorTron-3.9B. GatorTron was trained using clinical narratives and the HF dataset used only structured EHRs, therefore, there is no overlap between the HF dataset and the clinical narratives used to train GatorTron models.

**Experiment and Evaluation**

We compared feature density among patients using (1) the original ICD codes, (2) grouping using PheWAS diagnosis groups, and (3) subwords. The feature density is measured by the unique number of patients with the feature. Specifically, we counted the unique patients for each feature and ranked them from high to low for comparison. We randomly split the dataset into training, validation, and test using a ratio of 6:2:2. Table 2 provides the detailed numbers for the training, validation, and test sets. The ratios between the cases and non-cases among the training, validation, and test datasets are approximately 7:1. Therefore, this dataset is unbalanced.

**Table 2**. Training, validation, and test split.

| | **Training Set** | **Validation Set** | **Testing Set** |
|---|---|---|---|
| Encounters | 253,921 | 80,445 | 83,053 |
| Patients | 7,683 | 2,562 | 2,561 |
| Cases (with HF) | 987 | 313 | 302 |
| Non-case (no HF) | 6,696 | 2,249 | 2,259 |

For traditional machine learning models, we merged the validation with the training set and used a grid search and 5-fold cross validation to optimize the XGB and SVMs. For LSTM and LLMs, we followed the standard procedure to train models using the training set, identify the best models, stop the training using the validation performance on the validation set, and evaluate the performance using the test set. We conducted a study to examine feature combinations using demographics, diagnosis, and mediation for both T-LSTM and GatorTron-345M to identify the best feature combinations. Then, we used the best combination of features to train XGB, SVMs, T-LSTM, BERT, and GatorTron models and optimize the model parameters using the F1 score. A maximum of training epochs of 20 was used. The hidden layer size and fully connected layer size for T-LSTM were optimized as 128 and 64, respectively. Per optimization, the BERT and GatorTron-345M model used a learning rate of 1e-5, and the GatorTron-3.9B used a learning rate of 1e-6. We used evaluation metrics, including the area under the receiver operating characteristic curve (AUC), specificity, accuracy, precision, recall (recall equals sensitivity), and F1 score. F1 score was used as the primary evaluation metric for comparison. After identifying the best model, we applied it to the three cancer subgroups, including lung, breast, and colorectal, to examine the differences. The code for the paper is available at https://github.com/uf-hobi-informatics-lab/GatorTron-Prediction

**Results**

Figure 2 compares the three feature representation strategies using the original ICD codes, grouping by PheWAS groups, and subwords using the top 400 features. The original ICD codes have low density as they were designed for billing purposes. The PheWAS groups increased the density by aggregating patients with similar clinical conditions into the same group. The subword feature remarkably increased the density of the number of individuals per feature.

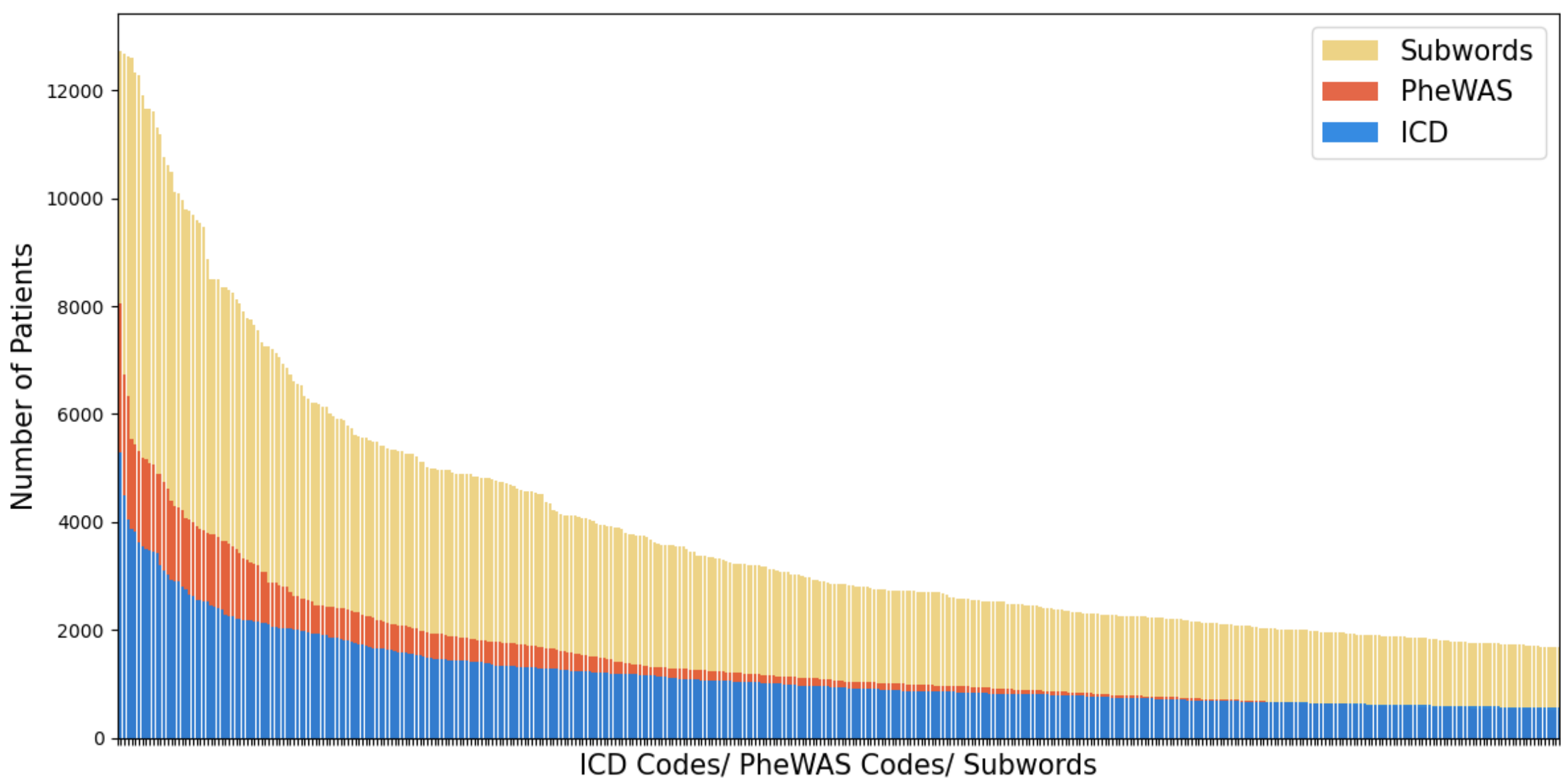

**Figure 2.** Comparison of feature density. The y-axis represents the number of unique people per feature, and the x-axis represents a distinctive feature based on the original ICD codes (blue), PheWAS groups (orange), and subwords (yellow).

**Table 3**. Feature combination for T-LSTM and GatorTron-345M.

| Models | Features | Feature encoding | F1 | Precision | Recall | AUC | Specificity | Accuracy |
|---|---|---|---|---|---|---|---|---|
| GatorTron-345M | Diagnosis | Subword | 0.605 | 0.705 | 0.530 | 0.848 | **0.970** | 0.918 |
| | Diagnosis+ Demographic | Subword | 0.610 | 0.695 | 0.543 | 0.918 | 0.968 | 0.918 |
| | Diagnosis+ Medication+ Demographic | Subword | **0.692** | **0.716** | **0.669** | **0.930** | 0.965 | **0.929** |
| T-LSTM | Diagnosis | Sequence | 0.491 | 0.479 | 0.503 | 0.815 | 0.927 | 0.877 |
| | Diagnosis+ Demographic | Sequence | 0.457 | 0.417 | 0.507 | 0.786 | 0.905 | 0.858 |
| | Diagnosis+ Medication+ Demographic | Sequence | 0.622 | 0.650 | 0.596 | 0.907 | 0.957 | 0.914 |

**Table 3** compares GatorTron-345M using subword features and T-LSTM using sequence features based on one-hot encoding. The F1 score was used as the primary evaluation score for optimization. Generally, both models achieved good performance when adding new information to the model, except for T-LSTM when adding demographics. Both models achieved the best F1 score using all features from diagnosis, medication, and demographics. GatorTron-345M outperformed the T-LSTM in all feature combinations. GatorTron-345M developed using diagnosis, medication, and demographic features achieved the best F1-score, precision, recall, and AUC of 0.692, 0.716, 0.669, and 0.930, respectively. **Table 4** compares GatorTron-3.9B and GatorTron-345M with BERT, T-LSTM, and traditional machine learning models, including XGB and SVMs. The T-LSTM model with sequence features outperformed traditional machine learning models using the one-hot feature. The LLMs (GatorTron and BERT) using subword features outperformed T-LSTM. The two GatorTron models achieved the best score on AUC, F1, precision, recall, and Accuracy, outperforming other models. GatorTron-3.9B achieved the best F1 score, which is used as this study's primary evaluation metric.

**Table 4**. Model comparison between 6 models.

| Model | Feature encoding | F1 | Precision | Recall | AUC | Specificity | Accuracy |
|---|---|---|---|---|---|---|---|
| XGB | One-hot | 0.290 | 0.592 | 0.192 | 0.825 | 0.982 | 0.889 |
| SVMs | One-hot | 0.302 | 0.376 | 0.252 | 0.756 | 0.944 | 0.863 |
| T-LSTM | Sequence | 0.622 | 0.650 | 0.596 | 0.907 | 0.957 | 0.914 |
| BERT | Subword | 0.643 | 0.631 | 0.656 | 0.890 | 0.949 | 0.914 |
| GatorTron-345M | Subword | 0.692 | 0.716 | 0.669 | 0.930 | 0.965 | 0.929 |
| GatorTron-3.9B | Subword | **0.699** | 0.684 | 0.715 | 0.896 | 0.956 | 0.927 |

**Table 5.** Subgroup analysis among lung, breast, and colorectal cancer using GatorTron_3.9B.

| Cancer Type | F1 | Precision | Recall | AUC | Specificity | Accuracy |
|---|---|---|---|---|---|---|
| Lung cancer (n = 940) | 0.634 | 0.774 | 0.537 | 0.872 | 0.977 | 0.920 |
| Breast cancer (n = 903) | 0.735 | 0.804 | 0.678 | 0.909 | 0.974 | 0. 935 |
| Colorectal cancer (n = 557) | 0.658 | 0.667 | 0.649 | 0.927 | 0.977 | 0.955 |
| More than one cancer (n = 161) | **0.824** | 0.750 | 0.913 | 0.957 | 0.949 | 0.944 |

We examined the performance differences among lung, breast, and colorectal cancer subgroups using GatorTron-3.9B. Table 5 provides the evaluation scores for the three cancer subgroups. The number of cases in each subgroup varies from 161 to 940. The performance varies across different cancer types. The comparison results show that GatorTron-3.9B achieved the best performance for predicting HF in patients with more than one cancer, with AUC, F1 scores, and recall of 0.957, 0.824, and 0.913 separately.

To understand how GatorTron models assess the risk of HF using narratives, we examined the attention weights assigned by GatorTron-3.9B to the subword features, which provides insight into the important narrative features driving the decision. We used LIME[32] (Local Interpretable Model-agnostic Explanations) package to visualize the important narrative features, where a darker orange color indicates important features. Figure 3 shows a breast cancer case with top highlighted keywords: 'abnormal electrocardiogram (ECG)', 'rosuvastatin', 'labetalol', and 'lisinopril'. These important narrative features indicate the presence of underlying cardiovascular disease potentially increasing the risk of cardiotoxicities in the future. Figure 4 shows a colorectal cancer case with 'abnormal ECG' and 'furosemide' highlighted. Abnormal ECG[33] is an indication of the potential of prior cardiac disease. Furosemide[34] is used to treat swelling that may be caused by heart failure.

occlusion and stenosis of left vertebral artery, weakness, shortness of breath, abnormal electrocardiogram [ ecg ] [ ekg ], encounter for fitting and adjustment of other gastrointestinal appliance and device, cerebral edema, cerebral infarction due to unspecified occlusion or stenosis of left middle cerebral artery, stroke nos, nan ; ascorbic acid / beta carotene / copper sulfate / selenite / vitamin e / zinc oxide, ascorbic acid 60 mg / beta carotene 5000 unt / copper sulfate 40 mg / dl - alpha tocopheryl acetate 30 unt / sodium selenite 0. 04 mg / zinc oxide 40 mg oral tablet, ascorbic acid, zinc oxide, vitamin e, beta carotene, copper sulfate, selenite, cephalexin, megestrol, rosuvastatin, iohexol, gadobutrol, sodium chloride, labetalol, atorvastatin calcium trihydrate, heparin, acetaminophen, bisacodyl, polyethylene glycol 3350, memantine, hydralazine, aluminum hydroxide, magnesium hydroxide, simethicone, atorvastatin, water, lidocaine, aspirin, clopidogrel, lisinopril,

Figure 3. Breast cancer case: visualization of the feature importance highlighted by GatorTron 3.9B.

nausea alone, flatulence, eructation, and gas pain, malignant neoplasm of colon, unspecified, atrial fibrillation, nonspecific abnormal electrocardiogram [ ecg ] [ ekg ] ; carbetapentane citrate / carbetapentane tannate / phenylephrine hydrochloride / phenylephrine tannate, epinephrine, bupivacaine, propofol, fentanyl, vecuronium, ephedrine, phenylephrine, glycopyrronium, neostigmine, glycopyrrolate, sodium gluconate, sodium acetate, magnesium chloride, lactate, calcium chloride, hydromorphone, hydroxyzine, alvimopan, heparin, tamsulosin, acetaminophen, albumin human, usp, albumin human - kjda, ondansetron, ondansetron hydrochloride anhydrous, simvastatin, cefoxitin, magnesium oxide, metoprolol succinate, potassium chloride, famotidine, lidocaine, naloxone, morphine, glucose, sodium phosphate, dibasic, sodium phosphate, monobasic, sodium phosphate, gabapentin, furosemide, sodium chloride, metoprolol, metoclopramide,

Figure 4. Colorectal cancer case: visualization of the feature importance highlighted by GatorTron 3.9B.

**Discussion and Conclusions**

Identifying cancer patients at risk of HF is critical to improving cancer treatment outcomes and safety. This study systematically examined traditional machine learning models using one-hot encoding features, T-LSTM using sequence encoding of structured codes, and LLMs using subword features using EHR data from UF Health. We identified a cancer cohort of 12,806 cancer patients, including 1,602 qualified cases who developed HF after cancer diagnosis and 11,204 non-cases without HF. The GatorTron-3.9B model using subword features achieved the best F1 scores, outperformed a machine learning model based on support vector machines by 39%, outperformed T-LSTM by 7%, and outperformed a widely used transformer model, BERT, by 5.6%. The analysis of feature density shows that subword features remarkably increased the feature density compared to other feature encoding strategies. This study demonstrates the efficiency of LLMs for disease risk prediction using subword features.

This study proposed a straightforward yet efficient strategy for LLMs using subword features derived from structured EHRs to predict the risk of diseases. Various feature representation strategies have been explored for disease risk prediction. Studies using the traditional machine learning models typically use the one-hot representation of structured EHRs, aggregating data points into a vector of zeros and ones, which is usually very sparse and could not capture the time-to-event structured medical codes. For example, there are 23,522 unique ICD-9 and ICD-10 codes and 19,685 clinical-level RxNorm drug codes in the original EHR data, which is very sparse among patients. This could be alleviated by grouping the codes using ontologies such as PheWAS disease groups. Later, deep learning models such as T-LSTM apply sequences of structured medical codes to capture the time sequence, which enriched the feature representation with time sequence. However, the feature sparsity issue still exists as the sequence features are based on the structured medical codes. The comparison results from Figure 2 show that the proposed subword features remarkably improved the feature density over the previous feature representation strategies. The experimental results show that GatorTron models with subword features outperformed traditional machine learning models using the one-hot encoding of structured codes and T-LSTM using the sequence of structured codes, supporting the effectiveness of using subword features in LLMs.

This study presents a potential solution to combine structured and unstructured EHRs for EHR-based predictions through automatic text summarization. Though LLMs have demonstrated good performance in NLP, there are no clear solutions to combine structured EHRs with unstructured EHRs in LLMs. Our experimental results show that LLMs with subword features are better surrogates than structured codes for disease risk prediction. Structured medical codes are easy to use for billing purposes but cannot reflect the differences between medical conditions related to each other. Grouping diseases using medical ontologies, such as the PheWAS disease group, could alleviate this issue, but it remains suboptimal. For example, PheWAS classified “Renovascular hypertension” into PheWAS group ‘401.3’ and classified “Pulmonary hypertension” into PheWAS group ‘415.2’, which were treated as two totally different diseases in traditional machine learning models and T-LSTM, but were treated as diseases closely related with each other owning to the word semantics captured by LLMs. Further aggregating the conditions using higher-level concepts such as the CCS groups could help, but this will cause over-aggregation issues that cannot differentiate two similar conditions. Using word semantics, LLMs can better capture the similarities and differentiate variations. For example, when considering the following two diagnoses, "Acute hepatitis B" and "Acute myocarditis", LLMs using subword features can capture the shared aspect of "acute" occurrence while distinguishing the differences between "hepatitis B" and "myocarditis."

This study has limitations. GatorTron models have a limitation of a maximum of 512 subwords; the subword features out of the 512 window were cut off, which may affect the performance of GatorTron models. We will explore solutions to increase the input window size through FlashAttention[35]. Breast cancer patients have an unbalanced gender distribution, with the majority of female patients. The black or African American cancer patients were slightly over-represented in the case group compared to the non-case group in our dataset, consistent with a higher risk for both of these conditions in this specific minority demographic. Future studies should examine new LLMs, such as the generative LLMs based on the GPT architecture, and explore automatic text summarization methods to integrate structured and unstructured EHRs for disease risk prediction.

**Acknowledgment**
This study was partially supported by a Patient-Centered Outcomes Research Institute® (PCORI®) Award (ME-2018C3-14754), the PARADIGM program awarded by the Advanced Research Projects Agency for Health (ARPA-H), grants from the National Cancer Institute, R01CA246418, R01CA246418-02S1, R21CA245858, R21CA245858-01A1S1, and R21CA253394-01A1, grants from the National Institute on Aging, NIA R56AG069880, R01AG080624, R01AG083039, R01AG080991, R01AG084236, R01AG084178, R01AG076234, and R33AG062884, National Institute of Allergy and Infectious Diseases, NIAID R01AI172875, the Cancer Informatics Shared Resource supported by the UF Health Cancer Center and the UF Clinical and Translational Science Institute Biomedical Informatics Program. The content is solely the responsibility of the authors and does not necessarily represent the official views of the funding institutions. We gratefully acknowledge the support of NVIDIA Corporation and the NIVIDA AI Technology Center (NVAITC) UF program. We would like to thank the UF Research Computing team, led by Dr. Erik Deumens, for providing computing power through UF HiperGator-AI cluster.